\documentclass[runningheads]{llncs}
\usepackage{graphicx}
\usepackage{amsmath,amssymb} 
\usepackage{color}

\begin{document}
\pagestyle{headings}
\mainmatter


\title{Appearance Fusion of Multiple Cues for Video Co-localization} 

\author{Koteswar Rao Jerripothula}
\institute{Indraprastha Institute of Information Technology Delhi (IIIT Delhi), India}

\maketitle

\begin{abstract}
This work addresses the joint object discovery problem in videos while utilizing multiple object-related cues. In contrast to the usual spatial fusion approach, a novel appearance fusion approach is presented here. Specifically, this paper proposes an effective fusion process of different GMMs derived from multiple cues into one GMM. Much the same as any fusion strategy, this approach also needs some guidance. The proposed method relies on reliability and consensus phenomenon for guidance. As a case study, we pursue the ``video co-localization" object discovery problem to propose our methodology. Our experiments on YouTube Objects and YouTube Co-localization datasets demonstrate that the proposed method of appearance fusion undoubtedly has an advantage over both the spatial fusion strategy and the current state-of-the-art video co-localization methods. 
\end{abstract}

\section{Introduction}
Unconstrained joint object discovery in a video collection is a challenging task. It is because the only supervision available is the association of other similar videos. Considering that there are only so many objects that the fully supervised algorithms (even deep-learning-based) can discover, such weak supervision becomes particularly essential. Apart from the commonness cue developed by the association, also known as co-saliency, there are also other cues from the video itself, such as saliency and motion cues. It is well known that different object-related cues are developed considering different goals, and integrating them is the goal of this paper. Uniquely, this paper explores the appearance domain for accomplishing such integration, instead of the usual spatial domain. As a case study, the problem of video co-localization (a kind of joint object discovery where the goal is to generate tight bounding box around the object jointly) is attempted in this paper, using such integration of multiple cues in the appearance domain. Essentially, this paper attempts to solve two problems: multi-cue integration and video co-localization.             

An object is a collection of multiple homogeneous regions, and a cue is a spatial probability distribution map. It is quite possible that a homogeneous region witnesses multiple cue values at different pixels. However, if a similar probability distribution is made available in the appearance (color) domain, a single color witnesses only one probability value. Therefore, one can say that the appearance domain can be banked upon to ensure uniformity in the homogeneous regions. Things become further uncertain when we have multiple cues. There are several problems in computer vision that require fusing multiple cues. One such problem is the video co-localization problem, which has not been explored much. Motivated by these observations, this paper ventures into the appearance domain for integrating the cues so that resultant cue can be useful in eventual video co-localization when combined with the spatiotemporal constraints.

\begin{figure}
    \begin{center}
        \includegraphics[width=1\linewidth]{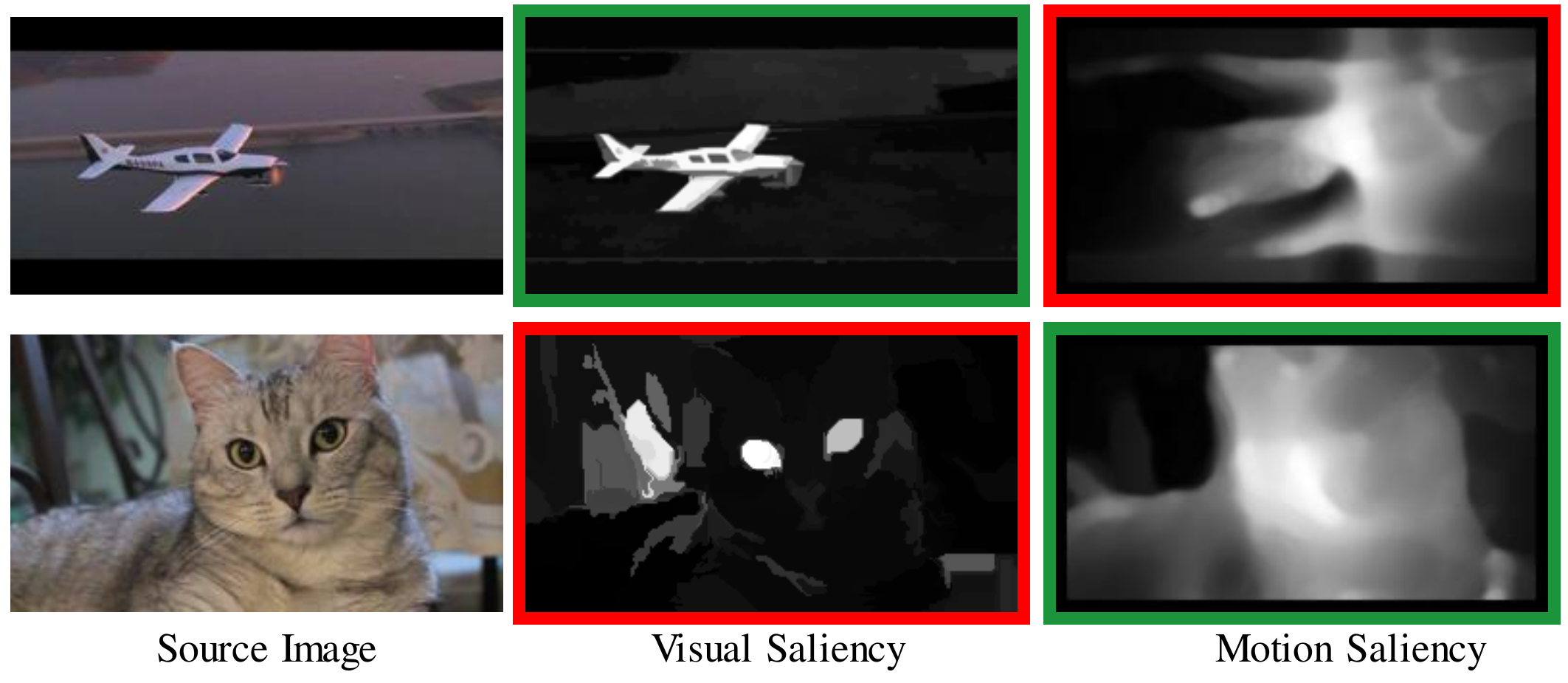}
    \end{center}
    
    \caption{Here are a few examples where the available cues are very inconsistent spatially. There are very few pixels which have similar likelihoods (foreground or background). The green bordered ones are readily good, whereas red ones are bad for object discovery. Why should we fuse good ones with bad another? Such scenarios rise the legitimate reliability issue while fusing multiple cues.}
    \label{fig:exmpl}
\end{figure}

There are quite a few exciting challenges in performing multi-cue integration and video co-localization. Intuitively, looking for consensus between the cues is the best way forward. However, if we see the examples in Fig.~\ref{fig:exmpl}, it is tough to build an algorithm that can successfully integrate (fuse) the visual saliency cue \cite{jiang2013salient} and the motion saliency cue \cite{6619174} just based on consensus. There is hardly any consensus in the two examples given. One may wonder if we should bother to fuse the good one (green) with the bad one (red)? Obviously, in such scenarios, we may end up spoiling the chances of good ones for accurate object discovery by fusing them with the bad ones. So, we require something more than consensus, and that is reliability. Thus, there is a need for incorporating reliability factor, too, in the algorithm we build. As far as video co-localization is concerned, developing the video co-saliency cue is challenging. Note that both the foreground similarity and the background variation are essential for building a robust co-saliency cue. Thus, there is a need for creating the right association to take full advantage of joint processing.

Previously, \cite{Jerripothula2016,kwak2015unsupervised} attempted the problem of co-localization in the spatial domain, and using the consensus factor only. In contrast, the proposed approach fuses different cues in the appearance domain while accounting for both consensus and reliability constraints. The idea is to first use GMM (Gaussian Mixtures Model) for modeling the appearance based on these cues, and then fuse these appearance models using the weights that depend on the consensus and reliability \cite{8269367} constraints. Notably, three cues have been used in this paper: co-saliency, visual saliency, and motion saliency. Several bounding box proposals are created when the fused appearance model yields the required fusion result of a rough object mask. From those proposals, an optimal one is selected via the proposed objective function, which respects both the masks developed and the spatiotemporal constraints across the video. Note that, different from previous works \cite{Jerripothula2016,kwak2015unsupervised}, which use the existing object proposals (limited by object categories seen during training), the proposals used here are just based on the rough object mask and the image edges. Such a mask-specific approach turns the proposed algorithm to be as generic as possible. Also, for developing the appropriate association for co-saliency generation, this paper extends the image co-saliency \cite{8269367} method to videos by applying it at multiple levels by building a hierarchy, instead of just at a single level. The hierarchical approach enables the algorithm to benefit from both very similar frames and very distinct frames.

The main contributions of this paper are twofold: 1) fusion of multiple cues in the appearance domain; 2) effective video co-localization with novel aspects in co-saliency generation and final localization.

\section{Related Works}
\subsection{Co-saliency}
Co-saliency detection has been extremely beneficial in various object discovery problems. For example, \cite{chang2011co} uses the co-saliency cue effectively in the co-segmentation problem \cite{li2018deep,chen2018semantic}. There have been previous attempts like in ~\cite{fu2013cluster} to fuse different cues, but they all fuse cues spatially. As far as videos are concerned, the term video co-saliency (different from video saliency\cite{8365810}) was introduced by \cite{7919200} to perform video co-segmentation while integrating different saliency cues. Other similar fusion approaches~\cite{8269367,7484309,7351686,7025663} fuse raw saliency maps of different images to generate co-saliency maps. All these techniques fuse spatially, whereas we attempt fusion in the appearance domain. Besides the contribution of appearance fusion, we also extend an image co-saliency technique to video through a hierarchical approach. To the best of our knowledge, only \cite{ACPR2017/JingLou,6675796} are related to hierarchy based co-saliency detection. While the hierarchy represented in \cite{ACPR2017/JingLou} depicts different scales of the image, the hierarchy in this paper depicts different levels at which the video frames of the dataset interact. Moreover, while \cite{6675796} uses hierarchical segmentation \cite{5557884} to obtain the co-saliency, we extend the existing image co-saliency idea for videos via a hierarchy.

\subsection{Co-localization}

Co-localization, similar to object co-segmentation, uses multiple images but to output bounding boxes. It was introduced by \cite{tang2014co}. Many of the image co-localization methods \cite{tang2014co,7457899,7298724} have been  later extended to video co-localization methods. For example, the image co-localization method \cite{tang2014co} was extended to the video co-localization method \cite{vcoleccv14}, where a quadratic optimization-based framework was proposed. The image co-localization method \cite{7457899} extends to the video co-localization method \cite{Jerripothula2016}. Then, an unsupervised image co-localization method \cite{7298724} extends to a video co-localization method \cite{kwak2015unsupervised}, which can localize the dominant objects even without necessarily requiring the video level labels. 

For video co-localization, \cite{vcoleccv14} and \cite{6248065} are the two frameworks that jointly locate common objects across all the videos. In \cite{vcoleccv14}, it used the quadratic programming framework to co-select bounding box proposals in all the video frames together. While in \cite{6248065}, it formed candidate tubes and co-selected tubes across the videos to locate the shared object. Recently, \cite{kwak2015unsupervised} proposed to develop foreground confidence for bounding boxes and select optimal ones while maintaining temporal consistency. However, it is a computationally intensive method requiring to match hundreds of proposals for a large number of frames of different videos. \cite{Jerripothula2016,8290832} propose an efficient way of video co-localization by introducing the concept of co-saliency activated tracklets. The idea is to develop co-saliency only for very few sampled frames, generate candidate tracklets in subsequent frames, and optimally localize the object in a video. However, such an approach may miss some significant changes that occur in the non-activators (non-sampled frames). We make up for this limitation by building a hierarchy on densely sampled frames. Moreover, we fuse appearance models instead of maps of cues, as performed in \cite{Jerripothula2016,8290832}. Several deep learning works like \cite{Tokmakov2016,rao2019common,leordeanu2020unsupervised,leordeanu2020coupling} report improved co-localization by using pre-trained deep learning networks. However, such approaches are limited to only the categories that the pre-trained network has seen during the initial training process. For comparing with such works, we use a pre-trained-network based saliency \cite{liu2019simple} extraction method in the proposed method.

\section{Proposed method}
\subsection{Overview}
The main objective here is to fuse GMMs resulting from different cues into a single GMM for effective video object discovery. Specifically, the video co-localization problem is pursued, which naturally has to deal with multiple cues like joint, saliency, and motion. Let's say we have joint ($P_j$), saliency (denoted as $P_s$) \cite{jiang2013salient} and motion (denoted as $P_m$) \cite{6619174} cues. Note that these cues have specific individual goals: (i) joint: to highlight common objects; (ii) saliency: to highlight different objects present; and (iii) motion: to highlight regions having unique motion. Due to the nature of their respective goals, different cues become effective in discovering objects in different scenarios depending upon the background types and the motion types, as shown in Table~\ref{tab:sce}. It is clear from the Table that there is some cue or other to facilitate object discovery in each scenario. In literature, we often find a spatial fusion of these cues, developing a spatial probability distribution for object discovery using such cues. In contrast, here, an appearance fusion strategy is proposed, developing appearance probability distribution, for object discovery ultimately. First, the proposed GMM fusion strategy is discussed. Then, different other components required in the video co-localization process are discussed.

\subsection{GMM Fusion}
Given a set of cues, the first task here is to develop some GMM components. It is done using the concept of tri-maps, where the important foreground and background seeds are identified. These seeds help in the development of GMM components that are required for our fused GMM. There is a slight difference between the proposed strategy and popular GrabCut strategy \cite{rother2004grabcut}. While our strategy uses multiple tri-maps, \cite{rother2004grabcut} uses only one tri-map. Then, we need to transform the fused GMM back to the spatial domain to carry out the object discovery problem. Essentially, the cues are being fused as GMMs in the appearance domain rather than the spatial domain. Now, each of these tasks will be discussed in detail.           

\begin{table}
    \centering
    \caption{Different cues become effective in different scenarios}\label{tab:sce}
     \vspace{1mm}

    \begin{tabular}{|c|c|c|}
    \hline
         & Simple Background & Complex Background\\
         \hline
        Unique Motion & Saliency, Joint and Motion & Joint and Motion \\
        General Motion & Saliency and Joint & Joint\\
        \hline
    \end{tabular}
\end{table}

\textbf{Generation of GMM components:}
Let the set of cues be denoted as $\mathcal{P}$. In the current problem context, $\mathcal{P}=\{P_k|k=c,v,m\}$. Let $I$ be the frame taken into consideration along with its pixel domain $D$.

In order to develop GMM components, we need tri-maps, which are developed in the following way. For the $k-th$ cue, let $T_k=\{T_k^f,T_k^b,T_k^u\}$ be the tri-map, where superscripts ($f,b,u$) denote foregound, background and unspecified regions of the tri-map and their corresponding labels in the tri-map are 1, -1 and 0, respectively. If $p$ denotes a pixel, it gets assigned to a particular region of $T_k$ based on the following criterion:

\begin{equation}
\begin{split}
p \in
\begin{cases}
T^f_k, & \text{if }P_k(p)>\Big(\phi_k+avg\big(P_k(P_k>\phi_k)\big)\Big);\\
T^b_k, & \text{if }P_k(p)<\phi_k;\\
T^u_k, & \text{otherwise};\\
\end{cases}\\
\end{split}
\end{equation}
where  $\phi_k$ denotes Otsu threshold value of $P_k$, and $avg$ denotes the average value. Note that by $P_k(P_k>\phi_k)$, we mean set all the values in $P_k$ that are greter than $\phi_k$. 

Each tri-map provides us corresponding sets of foreground seeds $T^f_k$ and the background seeds $T^b_k$. Each of these sets can give $N$ GMM components (typically, N=5, as suggested in the Grabcut strategy \cite{rother2004grabcut}) by clustering the set into $C$ groups. Let $C^f_k(c)$ and $C^b_k(c)$ denote $c-th$ foreground component of $k-th$ tri-map, respectively, where $c\in\{1,\cdots,N\}$. In this way, for forming the foreground fused GMM, we have total $|\mathcal{P}|*N$ components. The same is the case with the background fused GMM. Any GMM (foreground/background) component holds three parameters: weights ($\pi$), means ($\mu$) and co-variances ($\sigma$). While we keep the computation of $\mu$'s and $\sigma$'s same as in the Grabcut optimization \cite{rother2004grabcut}, we alter only the GMM weights to form our fused GMMs, as described next. 

\begin{figure}
    \begin{center}
        \includegraphics[width=1\linewidth]{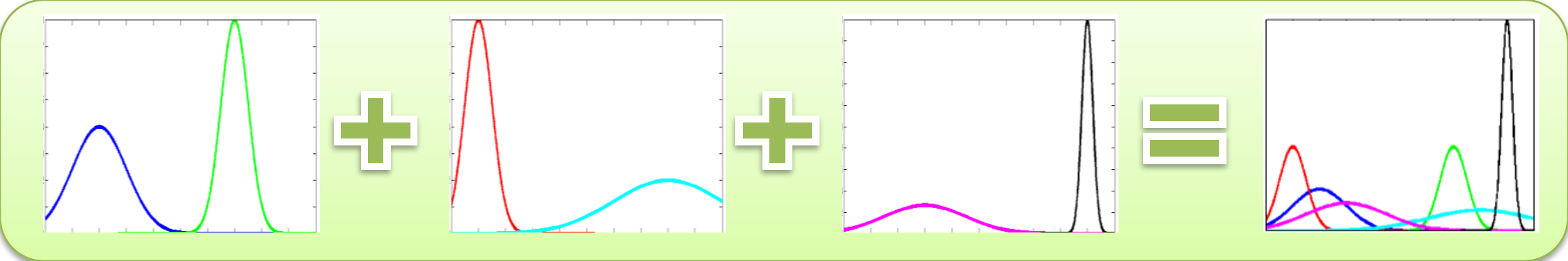}
    \end{center}
    
    \caption{A representation of our appearance fusion idea, where we stack the components of different appearance models into one appearance model. Note how the distribution parameters of different components (means and variances) just get transferred here. Similarly, our GMM components' distributions parameters also just get transferred to create the new appearance model.}
    \label{fig:gmmmix}
\end{figure}

\textbf{Generation of fused GMM:} Let us now discuss how to weight our GMM components appropriately. Weights signify the importance of a component. In the Grabcut strategy, the components' weights are computed as fractions of the seeds that belonged to the components. However, in our case, a pixel may serve as a seed more than once, for there are multiple tri-maps. As a result, we design our weights such that the components that have good reliability and good consensus get higher weights than unreliable ones and with lower consensus.

For this purpose, we utilize the map called consensus aware reliability map ($X$) developed by us. We develop the consensus aware reliability map ($X$) using
\begin{equation}\label{eq:x}
X(p)=\sum\limits_{k=1}^{|\mathcal{P}|}{T_k(p)\times\psi(P_k)}
\end{equation}
where the signed sum of reliability scores $\psi(\cdot)$ is being performed. By the signed sum, we mean coefficients could be 1, -1, or 0 depending upon pixel's value in a particular tri-map. Essentially, if the signs are similar for a particular pixel, there is a consensus. We design the equation such that: if the consensus is high, the score will be as high as possible, and if the consensus is poor, the score will become as low as possible. Note that the reliability score $\psi(\cdot)$ is computed using \cite{8269367} for a given cue. \cite{8269367} computes these reliability scores according to the overlap between Gaussian fits of foreground and background distributions and the foreground concentration. The labels $T_k(p)\in \{1,-1,0\}$ of respective tri-maps form coefficients of these scores. Such design brings out both the consensus concept and reliability quite nicely: if they are reliable and have consensus with each other in terms of their known labels, the score will be high. By known labels, we mean foreground (1) and background (-1) labels. A total of $3^{|\mathcal{P}|}$ cases are possible if $|\mathcal{P}|$ cues are available. 

The weight of any foreground GMM component, say $\pi(C^f_k(c))$, is determined as given below:
\begin{equation}
\pi(C^f_k(c))=0.5*\Big(1+\frac{\sum_{p\in C^f_k(c)}X(p)}{|C^f_k(c)|}|\Big)
\end{equation}, where the average score of pixels belonging to $C^f_k(c)$ is adjusted to 0-1 range, to reflect foreground confidence. Similarly, in the case of background GMM component, the weight computation is as follows:
\begin{equation}
\pi(C^b_k(c))=0.5*\Big(1-\frac{\sum_{p\in C^b_k(c)}X(p)}{|C^b_k(c)|}\Big)
\end{equation}, where the average score of the pixels belonging to $C^f_k(c)$ is negated and adjusted to 0 to 1 range to reflect background confidence. Note that since these weights need to sum to 1, we normalize these weights accordingly. With GMM components' weights set right, we now have one fused foregfound GMM and one fused Background GMM, which can be used in the GrabCut energy function for segmentation. We just perform one iteration of GrabCut to obtain the required rough object mask with the common backgrounds in the tri-maps as fixed background. 

\subsection{Video Co-localization}
As mentioned in the overview of this section, we require three cues for video co-localization. Out of these three, the saliency cue ($P_s$) and motion saliency cue ($P_m$) are extracted using the already existing \cite{jiang2013salient} and \cite{6619174} works, respectively. However, for co-saliency cue ($P_c$), we modify the existing image co-saliency \cite{8269367} work for it to work as a video co-saliency detection algorithm. We will now discuss how we generate this video co-saliency cue for providing it to the GMM fusion (discussed above) and how exactly we perform video co-localization once we have the rough object mask developed using the GMM fusion.   

\textbf{Co-saliency Cue Generation:} We develop a hierarchy for generating co-saliency cues. The hierarchy of video frames provides us a multi-level representation of different relationships existing in the dataset. Such a representation facilitates the strategic exploitation of multi-level co-saliency maps to generate a final one. Denote such a dataset as $\mathcal{V}=\{V_1,V_2,...,V_n\}$ having total $n$ videos. Let each video be denoted as $V_i=\{F_{i}^1,F_{i}^2,.....,F_{i}^{|V_i|}\}$, i.e., set of the comprising frames.

\begin{figure}
    \begin{center}
        \includegraphics[width=1\linewidth]{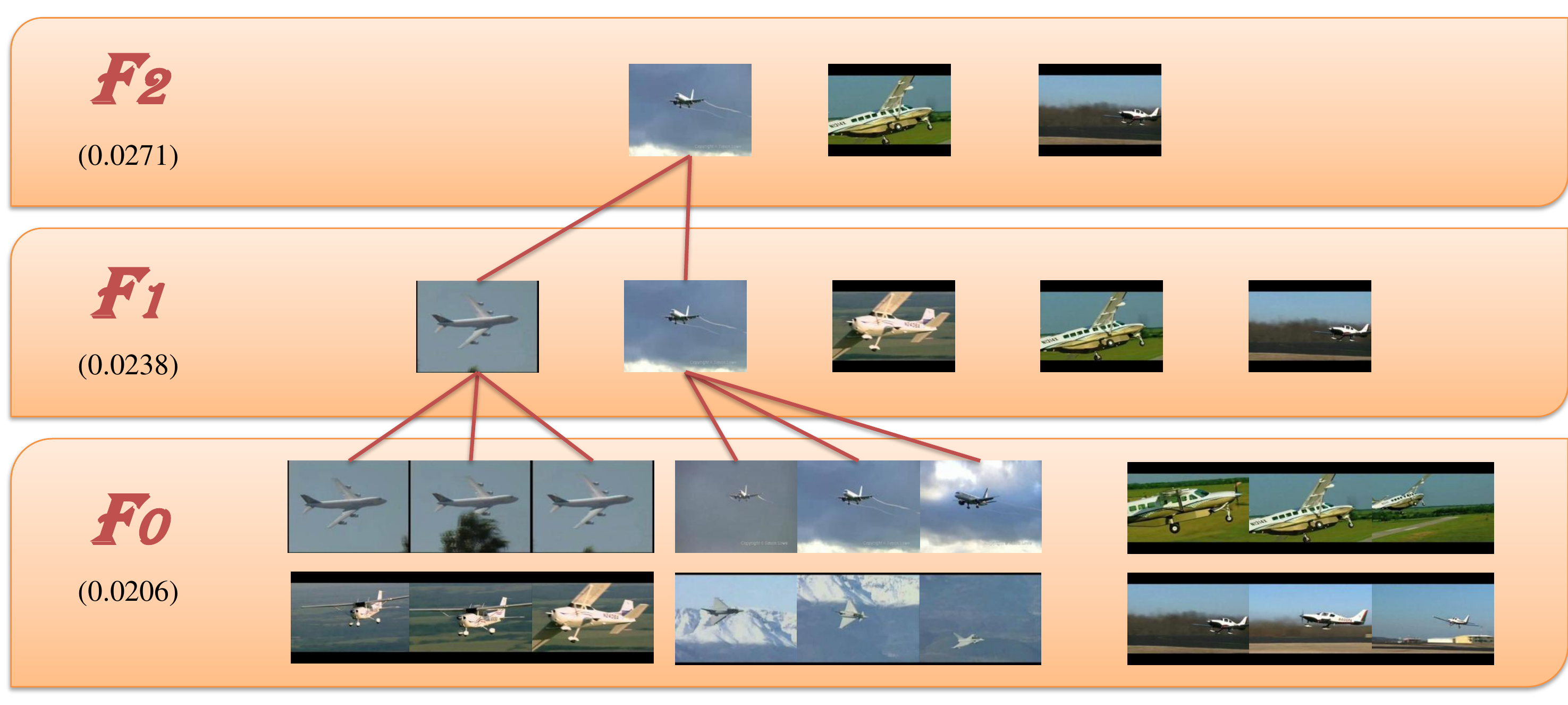}
    \end{center}
    
    \caption{An illustration of the hierarchy development: A higher level comprises representatives of frames at a lower level. The variation at any level increases from bottom to top, suggesting the potential to benefit from multiple levels of variation.}
    \label{fig:hei}
\end{figure}

We build our hierarchy of frames in the following manner: At the ground level (denoted as $\mathcal{F}_0$), all the video frames are present. At a level higher than $\mathcal{F}_0$ is $\mathcal{F}_1$, where the representatives of $\mathcal{F}_0$ are present, as determined by the k-means clustering. Similarly, $\mathcal{F}_2$ consists of representatives of $\mathcal{F}_1$, and so on.  Note in Fig. \ref{fig:hei} how GIST~\cite{oliva2001modeling} (a global descriptor) variation (in brackets) keeps increasing as we go to the higher level in the hierarchy of representative frames. The GIST variation here means variance of GIST features of the frames present on a particular hierarchy level. We choose frames nearest to the k-means cluster centers as the representative frames. We adjust the number of clusters in a manner that we ensure there are 15 frames on an average in a cluster. The hierarchy building is terminated when the variation is above a certain threshold, or the number of representatives is less than a minimum number.

Once we generate the hierarchy, we can now use an image co-saliency technique in \cite{8269367} to create co-saliency maps at multiple levels using the frames available at the lower levels. Now, the idea behind image co-saliency in \cite{8269367} is to warp saliency maps of other images to the representative image for obtaining a generalized co-saliency for the representative image. Then, the generalized co-saliency maps are propagated to other images by warping them back to generate co-saliency maps of others. \cite{8269367} used only a two-level hierarchy. In our case, since we have multiple levels, we start from the lowest level and generate co-saliency maps at each level progressively using the co-saliency maps of the lower level, up to the highest one. The idea is to witness the varying degrees of similarity and variation at multiple levels, and benefit from all, not just the bottom-most level ($F_0$), a kind of nearest neighborhood approach.  

Now, with co-saliency maps available at multiple levels (except $F_0$), as far as propagation is concerned, we start disseminating from the highest level generating a new co-saliency map at each level for ultimately yielding the co-saliency maps of $F_0$. For this propagation step, we set the weights for the fusion of disseminating warped saliency maps and the current co-saliency map (saliency map for $F_0$) to the number of levels covered so far while propagating and one, respectively. Such a weight-assignment appropriately acknowledges the importance of disseminating warped co-saliency maps over the current one. In this way, we can eventually generate final co-saliency maps at $F_0$ level as well, which had only saliency maps at the beginning. 

In this way, in our hierarchy based co-saliency approach, we have extended the \cite{8269367} co-saliency method for its application on video frames by applying it at multiple levels of our hierarchy instead of just a single level.

\begin{figure*}
    \begin{center}
        \includegraphics[width=1\linewidth]{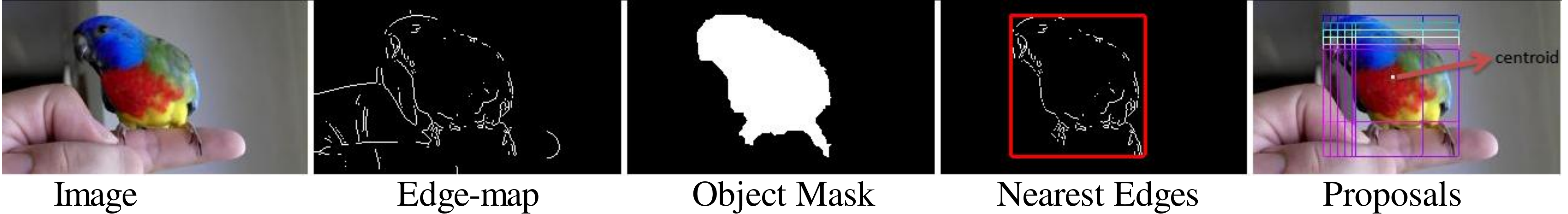}
        
    \end{center}
    
    \caption{Generation of bounding box proposals: (i) Using distance transform of the original edge map, the nearest edges of the rough object mask's foreground pixels are obtained. (ii) A reference bounding box (in red) is created using extreme nearest edges. (iii) Several proposals are generated using sampled nearest edges such that the centroid of the foreground segment remain inside the proposal.}
    \label{fig:bbgen}
\end{figure*}

\textbf{Bounding-Box Proposal Generation:}
To generate mask-specific proposals, we first try to build a reference bounding-box using the rough object mask obtained after appearance fusion and the edge pixels present in the image. For this purpose, we identify the nearest edge pixels (in the edge map) of foreground pixels (of our rough object mask) as the required edge pixels. Specifically, we create distance-transform of the edge map and then find the nearest edges for our mask's foreground pixels. We use \emph{vl\_imdisttf} function in vlfeat \cite{vedaldi08vlfeat} library for this purpose. In this way, a new mask-specific edge map is developed. A reference bounding box is generated over the extreme edge pixels in such a map, as shown in the second-last image of Fig.~\ref{fig:bbgen}. This reference bounding box helps us score candidate bounding box proposals generated from the edge pixels available in the mask-specific edge map. We consider only those bounding boxes as candidates that contain centroid of our foreground pixels within them and also pass through the sampled edge pixels. Such requirements ensure the generation of limited candidates while sufficiently respecting our rough object mask. Denote $\mathcal{B}^j_i$ as a set of all such proposals in the frame $F^j_i$.

\begin{figure}
    \begin{center}
        \includegraphics[width=1\linewidth]{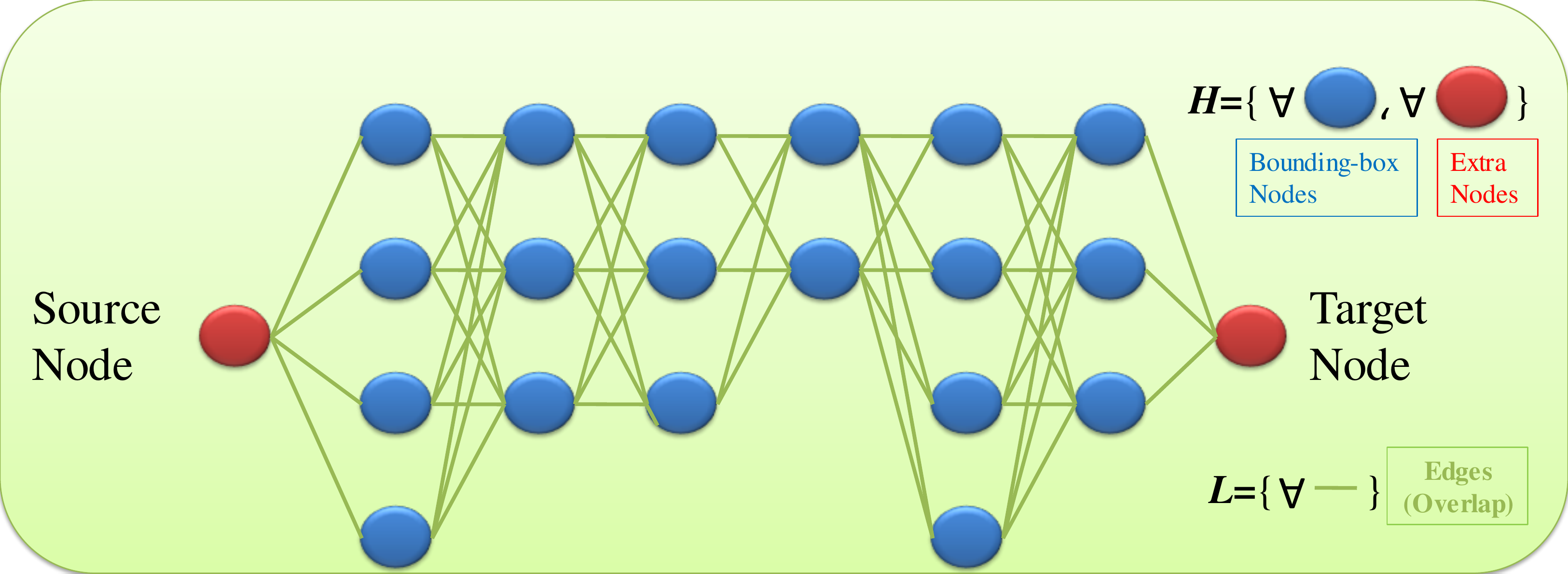}
    \end{center}
    
    \caption{Graph to represent the bounding boxes as nodes and overlaps as linkages.}
    \label{fig:grp}
\end{figure}

\textbf{Graph:} For any video, we construct a directed graph $\mathcal{G}<H, L>$ which connects our mask-specific bounding box proposals across adjacent frames as shown in the Fig.~\ref{fig:grp}, where $ H $ denotes a set of all the bounding box proposals as nodes, and $ L $ denotes the set of links that connect the nodes. For simplifying our notations, let $ H $ and $ L $ also denote the sets of weights of nodes and edges. The weights of the nodes (proposals) are defined as the perimeters of the proposals. The weights of the links are defined as Euclidean distance between the proposals they connect in terms of top, bottom, left, and right coordinates. Since the highest perimeter is possible only for our reference bounding box, the higher the perimeter means closer is the proposal to reference. Moreover, more the Euclidean distance between proposals from adjacent frames, lesser is their overlap. We also add source and target nodes at the two ends of the graph to facilitate the formulation of the shortest path problem to select optimal proposals for localization of the objects.

\textbf{Objective Function:} The goal of the shortest path problem is to traverse through a path from source to target at the lowest possible cost. Let us denote the node selection variable and the linkage selection variable as $\mathbf{z}$ and $\mathbf{y}$, respectively. Since the selection of links inherently takes care of the bounding boxes' selection, we formulate our objective function as

\begin{equation} \label{eq:obj}
\begin{aligned}
&\min_\mathbf{y,z}  &   & \sum_{(m,n)\in \mathcal{L}}{\mathbf{y}_{mn}\big(-\log(H_mH_n)+\lambda L_{mn}\big)}\\
&\text{s. t.} &  &\sum_{k=1}^{|\mathcal{B}^j_i|} \mathbf{z}(j,k)=1, \ \forall F_{i}^j \in V_i\\
&\text{     } &  & \mathbf{y}_{mn}=\mathbf{z}_m=\mathbf{z}_n\\
&\text{     } &  & \mathbf{y}\in\{0,1\},\mathbf{z}\in\{0,1\}\\
\end{aligned}
\end{equation}
where $\mathbf{y}_{mn}\in \mathbf{y}$ indicates whether the edge between the nodes $m$ and $n$ (consequently the nodes also) will be part of the shortest path or not. For developing the total cost of path, we need to account for both the nodes and the linkages we come across while traversing. The cost ($-\log(H_mH_n)$)for selecting a node-pair which has been computed as negative logarithm of the product of the two node weights encourages legitimate selection of the proposals with higher perimeter, signifying reliance on our rough object mask. The cost ($L_{mn}$) for selecting a linkage encourages the linkages having higher overlap between the the proposals across the linkages, ensuring a spatiotemporally smooth tube. The two costs are traded off by a parameter $\lambda$. Moreover, the objective function is subjected to the requirement of selecting one candidate bounding box per frame. Another requirement is that if a linkage is selected, nodes across them must also be selected. This objective function can be easily solved using Dijkstra's shortest path algorithm to compute $\mathbf{z}$, which gives the list of the proposals selected for localizing the object in each frame of the video $V_i$. 


\section{Experiments}

\subsection{Experimental Setup}
We use the YouTube Objects and YouTube Co-localization datasets, which consist of 10 categories belonging to either animal (e.g., cow and cat) or vehicle (e.g., train and car). Each category has significant intra-class variation, making it a challenging dataset for video co-localization problem. Following the literature, we use the CorLoc evaluation metric, i.e., the percentage of frames that satisfy the IoU (Intersection over Union)$>0.5$ condition. Note that while reporting the results of different methods, we report the best result in red and the second-best result in blue. We set our balancing parameter $\lambda$ as 5 in all our experiments. 

\begin{figure*}
    \begin{center}
        \includegraphics[width=1\linewidth]{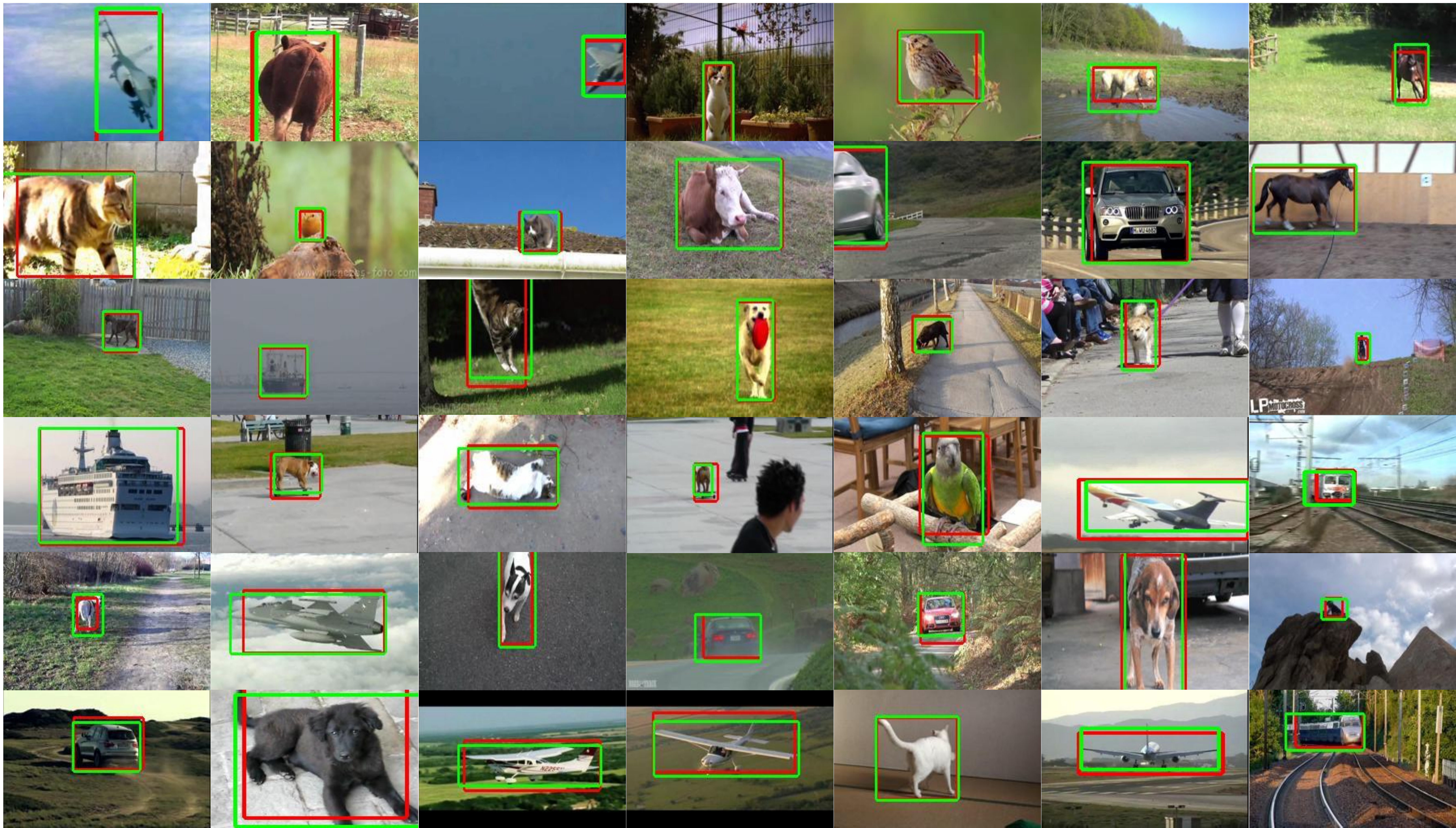}
    \end{center}
    
    \caption{Sample object localization results (red) in a wide variety of video frames across the YouTube Objects dataset along with their corresponding ground truth bounding boxes (green). It can be seen that they are quite close.}
    \label{fig:inter}
        \end{figure*}

\begin{table*}
    \begin{center}
        \caption{CorLoc results on YouTube-Objects dataset under weakly supervised scenario (with video level labels).} \label{tab:vcol}

        \begin{tabular}{|l||c|c|c|c|c|c|c|c|c|c||c||c|}
            \hline
            & aeroplane & bird & boat & car  & cat  & cow  & dog  & horse & bike & train & avg & std \\
            \hline
            Prest et al.~\cite{6248065}                      & 51.7      & 17.5 & 34.4 & 34.7 & 22.3 & 17.9 & 13.5 & 26.7  & 41.2      & 25.0  & 28.5 & 11.9 \\
            Joulin et al.~\cite{vcoleccv14}                   & 25.1      & 31.2 & 27.8 & 38.5 & 41.2 & 28.4 & 33.9 & 35.6  & 23.1      & 25.0  & 31.0 & \textbf{\textcolor{red}{6.1}} \\
            Kwak et al.~\cite{kwak2015unsupervised}& 56.5      & 66.4 & 58.0 & \textbf{\textcolor{blue}{76.8}} & 39.9 & 69.3 & 50.4 & 56.3  & 53.0      & 31.0  & 55.7 & 13.5\\
            Jerri. et al.~\cite{Jerripothula2016}                         & 65.7      & 59.6 & 66.7 & 72.3 & 55.6 & 64.6 & 66.0 & 50.4  & 39.0      & \textbf{\textcolor{blue}{42.2}}  & 58.2 & 11.2\\
            Jerri2.et al.~\cite{8290832}&\textbf{\textcolor{red}{76.8}}&\textbf{\textcolor{red}{68.3}}&66.7&64.4&50.4&66.9&69.5&\textbf{\textcolor{blue}{57.4}}&41.0&38.8&60.0& 12.7\\
            Sharma ~\cite{sharma2018foreground}&-&-&-&-&-&-&-&-&-&-&54.0&-\\
            Rochan et al.~\cite{rochan2015weakly}&56.0&30.1&39.7&\textbf{\textcolor{red}{85.7}}&24.8&\textbf{\textcolor{red}{87.8}}&55.7&\textbf{\textcolor{red}{60.3}}&61.8&\textbf{\textcolor{red}{51.8}}&55.4&20.7\\

          Jun Koh~\cite{jun2016pod}&64.3&63.2&\textbf{\textcolor{red}{73.3}}&68.9&44.4&62.5&\textbf{\textcolor{red}{71.4}}&52.3&\textbf{\textcolor{red}{78.6}}&23.1&\textbf{\textcolor{blue}{60.2}}& 16.4\\
            \hline
            Avg. ($P_v, P_m$) & 55.9& 46.2 & 65.2 & 57.1& 43.6&46.5 &    62.4       &      50.4     &53.0 &27.6 &50.8& 10.8\\
            Avg. ($P_c, P_v, P_m$) & 63.7&  57.7& 63.8 & 64.3& 48.1&52.0 &  66.7         &     48.8      & \textbf{\textcolor{blue}{55.0}}& 29.3&54.9&11.3\\
            Appearnce Fusion & 69.9& 61.5& 63.0 & 68.8&\textbf{\textcolor{blue}{57.9}} & 68.5&   69.5        & 56.6         &54.0 & 39.7&60.9&9.5\\
            Proposed Method& \textbf{\textcolor{blue}{70.9}} & \textbf{\textcolor{blue}{67.3}}          & \textbf{\textcolor{blue}{72.5}} & 75.0          & \textbf{\textcolor{red}{59.4}} & \textbf{\textcolor{blue}{73.2}} & \textbf{\textcolor{blue}{70.9}}          & \textbf{\textcolor{blue}{57.4}}          & \textbf{\textcolor{blue}{55.0}}          & \textbf{\textcolor{red}{51.8}} & \textbf{\textcolor{red}{65.3}} & \textbf{\textcolor{blue}{8.6}}\\
            \hline
        \end{tabular}
    \end{center}
       \end{table*}

\subsection{Video Co-localization Results}

\textbf{Weakly Supervised Scenario:} In the weakly-supervised setup, we build our hierarchy upon the videos of the same category. In Table \ref{tab:vcol}, to demonstrate the importance of each of the proposed method's components, we report four types of results on YouTube Objects Dataset. First, Avg. ($P_v, P_m$) denotes results obtained by averaging $P_v$ and $P_m$ (visual and motion saliency cues); it is a baseline. Second, Avg. ($P_c, P_v, P_m$) denotes results obtained when we add the developed $P_c$ to the first one. Third, we report the results obtained by the proposed appearance fusion scheme instead of averaging. In all these three types, we use a tight bounding box across the largest foreground component of a rough object mask for obtaining the results. Fourth, we report results obtained by our full method, where we also consider the spatiotemporal constraints. 

\begin{figure*}
    \begin{center}
        \includegraphics[width=1\linewidth]{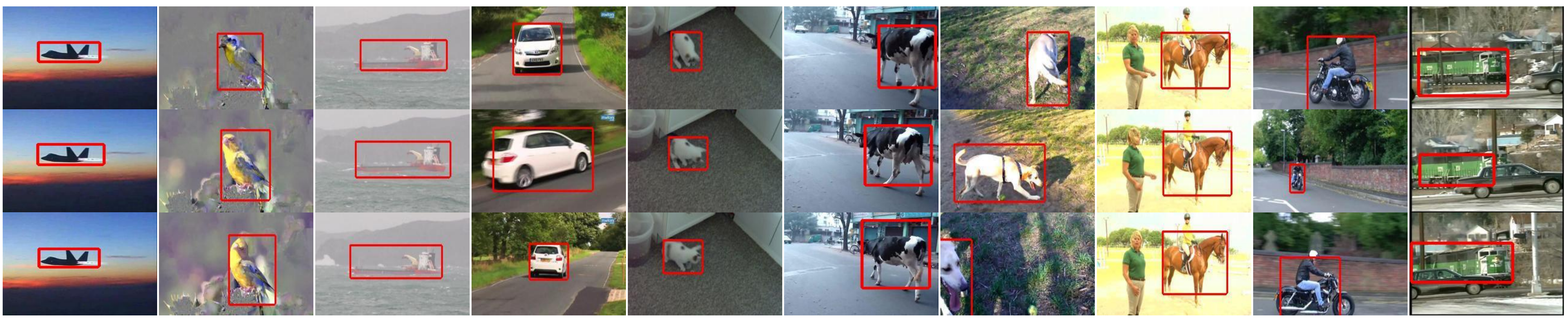}
    \end{center}
    
    \caption{Sample object co-localization results in different videos. The proposed method can accommodate variations of size, location, view, and so on.}    \label{fig:intra2}
    \end{figure*}

We can note how each of our components progressively contributed: (i) upon introduction of $P_c$, the performance improves by 4.1\%; (ii) upon introduction of appearance fusion, the performance improves by 6\%; and upon introduction of SPL (i.e., the proposed method), there is further improvement of 4.4\%. It can also be seen in this Table that our method outperforms several existing state-of-the-art methods comfortably. We achieve 8.5\% relative improvement over \cite{jun2016pod} (the best among existing ones) in terms of the average (avg) of CorLoc scores across the categories. We also report the standard deviation (std) across the categories to evaluate robustness, and our method obtained the second-best standard deviation. These results show that the proposed method is not just accurate but robust as well. As far as our qualitative results are concerned, while Fig.~\ref{fig:inter} demonstrates the closeness of our results (red) to the ground truths (green), the Fig.~\ref{fig:intra2} demonstrates the ability of our method to handle all kinds of variations within the video. Additionally, in the supplementary material to this document, we provide our sample video results.
    
    \begin{table*}
        \begin{center}
        \caption{CorLoc results on YouTube-Objects dataset using pre-trained networks. Here, we replace saliency extraction method \cite{jiang2013salient} with \cite{liu2019simple}, which uses VGG as backbone.} \label{tab:svcol}
        \setlength{\tabcolsep}{0.5 pt}
        \begin{tabular}{|l||c|c|c|c|c|c|c|c|c|c||c|}
      
            \hline
            & aeroplane     & bird          & boat          & car           & cow           & cat           & dog           & horse         & bike     & train         & avg            \\\hline
           Rao et al.~\cite{rao2019common}&44.3&68.6&56.7&63.5&50.0&70.7&71.2&\textbf{\textcolor{red}{75.9}}&73.8&55.5&63.0\\   
        Tokmakov et al.\cite{Tokmakov2016} & 76.1&57.7&\textbf{\textcolor{blue}{77.7}}&68.8&\textbf{\textcolor{blue}{71.6}}&\textbf{\textcolor{blue}{75.6}}&\textbf{\textcolor{blue}{87.9}}&71.9&\textbf{\textcolor{blue}{80.0}}&52.6&\textbf{\textcolor{blue}{72.0}}\\
            Leordeanu ~\cite{leordeanu2020unsupervised}& \textbf{\textcolor{blue}{87.4}}&72.7&77.2&64.6&62.4&75.0&82.7&56.7&52.9&39.5&67.1\\
            Leordeanu2 ~\cite{leordeanu2020coupling}&\textbf{\textcolor{red}{88.2}}&\textbf{\textcolor{red}{82.5}}&62.7&\textbf{\textcolor{blue}{76.7}}&70.9&50.0&81.9&51.8&\textbf{\textcolor{red}{86.2}}&\textbf{\textcolor{blue}{55.8}}&70.7\\
            \hline
            Ours (using \cite{liu2019simple})&81.6 & \textbf{\textcolor{blue}{78.9}} & \textbf{\textcolor{red}{85.5}} & \textbf{\textcolor{red}{91.1}}& \textbf{\textcolor{red}{77.4}} &\textbf{\textcolor{red}{88.2}} &  \textbf{\textcolor{red}{89.4}}         &  \textbf{\textcolor{blue}{74.4}}         &78.0 & \textbf{\textcolor{red}{60.2}}&\textbf{\textcolor{red}{80.5}}\\
            \hline
        \end{tabular}
    \end{center}
       \end{table*}

Recently, several deep-learning-based co-localization results have been reported. Since this is a heuristic approach, in Table \ref{tab:svcol}, we use a pre-trained-network based saliency (PoolNet \cite{liu2019simple} with VGG as its backbone pre-trained network) extraction method in the proposed method to compare with these methods, which also use such pre-trained networks. Our full method obtains nearly 13\% relative improvement. 

Note that YouTube Objects Dataset has only about 1.4k annotations (just one frame per video shot). In contrast, YouTube Co-localization Dataset has about 15k annotations (at every 10th frame of nearly 1.2k videos). As presented in Table ~\ref{tab:cvcol}, the proposed method outperforms state-of-the-art video co-localization method\cite{8290832} of this dataset as well. In turn out that just our appearance fusion is enough to outperform this particular state-of-the-art.

\begin{table*}
    \begin{center}
        \caption{CorLoc results on YouTube Co-localization dataset in weakly-supervised scenario (with labels).} \label{tab:cvcol}
        \begin{tabular}{|l||c|c|c|c|c|c|c|c|c|c||c|}
      
            \hline
            & aeroplane     & bird          & boat          & car           & cow           & cat           & dog           & horse         & bike     & train         & avg            \\\hline

        Jerri2 et al.\cite{8290832} & \textbf{\textcolor{blue}{61.5}}&52.5&64.3&\textbf{\textcolor{blue}{62.5}}&47.3&\textbf{\textcolor{blue}{69.9}}&55.6&53.9&45.5&34.3&54.7\\

            \hline
            Appearance Fusion &61.1& \textbf{\textcolor{blue}{60.8}} & \textbf{\textcolor{blue}{69.0}} & 59.5&\textbf{\textcolor{blue}{51.2}} &68.6 & \textbf{\textcolor{blue}{61.9}}&  \textbf{\textcolor{blue}{56.8}}         &\textbf{\textcolor{red}{57.0}} &\textbf{\textcolor{blue}{36.2}}&\textbf{\textcolor{blue}{58.2}}\\
            Proposed Method &\textbf{\textcolor{red}{67.2}} & \textbf{\textcolor{red}{67.9}} & \textbf{\textcolor{red}{70.4}} & \textbf{\textcolor{red}{62.8}}&\textbf{\textcolor{red}{57.4}} &\textbf{\textcolor{red}{74.8}} &  \textbf{\textcolor{red}{67.3}}         &  \textbf{\textcolor{red}{60.6}}         &\textbf{\textcolor{blue}{52.9}} & \textbf{\textcolor{red}{47.0}}&\textbf{\textcolor{red}{62.8}}\\
            \hline
        \end{tabular}
    \end{center}
    \end{table*}
        
    \begin{table*}
    \begin{center}
        \caption{CorLoc results on YouTube-Objects dataset in unsupervised scenario (without video level labels).} \label{tab:uvcol}
        \begin{tabular}{|l||c|c|c|c|c|c|c|c|c|c||c|}
            \hline
            & aeroplane     & bird          & boat          & car           & cow           & cat           & dog           & horse         & bike     & train         & avg\\
            \hline
                        Haller~\cite{haller2017unsupervised}&\textbf{\textcolor{blue}{76.3}}&\textbf{\textcolor{red}{71.4}}&65.0&58.9&\textbf{\textcolor{red}{68.0}}&55.9&70.6&33.3&\textbf{\textcolor{red}{69.7}}&42.4&61.1\\
            Kwak et al.~\cite{kwak2015unsupervised}     & 55.2          & 58.7          & 53.6          & \textbf{\textcolor{blue}{72.3}} & 33.1          & 58.3          & 52.5          & 50.8 & 45.0 & 19.8          & 49.9        \\
            Jerri. et al.~\cite{Jerripothula2016} & 66.7 & 48.1          & 62.3 & 51.8          & 49.6 & 60.6 & 58.9          & 41.9          & 28.0          & \textbf{\textcolor{blue}{47.4}} & 51.5\\
            Jerri2.et al.~\cite{8290832}&75.2&67.3&66.7&62.5&50.4&66.9&67.4&56.6&36.0&40.5&58.9\\
            Croitoru et al.~\cite{croitoru2017unsupervised}&\textbf{\textcolor{red}{77.0}}&\textbf{\textcolor{blue}{67.5}}&\textbf{\textcolor{red}{77.2}}&68.4&\textbf{\textcolor{blue}{54.5}}&\textbf{\textcolor{blue}{68.3}}&\textbf{\textcolor{blue}{72.0}}&\textbf{\textcolor{blue}{56.7}}&44.1&34.9&\textbf{\textcolor{blue}{62.1}}\\\hline
            Ours (U)&70.3 & 63.5 & \textbf{\textcolor{blue}{70.3}} & \textbf{\textcolor{red}{74.1}}&52.6 &\textbf{\textcolor{red}{70.1}} &  \textbf{\textcolor{red}{73.1}}         &  \textbf{\textcolor{red}{58.9}}         &\textbf{\textcolor{blue}{57.0}} & \textbf{\textcolor{red}{50.0}}&\textbf{\textcolor{red}{64.0}}\\
            \hline
        \end{tabular}
    \end{center}

\end{table*}
\textbf{Unsupervised Scenario:} In the weakly supervised setup, we build our hierarchy upon any video collection,i.e., there is no requirement of any video level label. In such a mode, the algorithm has to entirely rely on its neighborhood building technique for jointly localizing the objects instead of human annotations of video-level labels of semantic category. The results are given in Table~\ref{tab:uvcol}. We obtain nearly 3\% relative improvement over \cite{croitoru2017unsupervised} (current stat-of-the-art) in terms of the average of CorLoc scores across the categories. Note that after neglecting the video-level labels of the semantic category in the unsupervised mode, our performance slightly drops from 65.3 to 64, which is just a drop of just 1.3. Given such a less drop when there is such an increase in diversity, our hierarchy-based neighborhood approach seems to provide excellent neighborhood exposure for object discovery and saves human efforts in providing video-level labels. 
\begin{figure*}
    \begin{center}
        \includegraphics[width=1\linewidth]{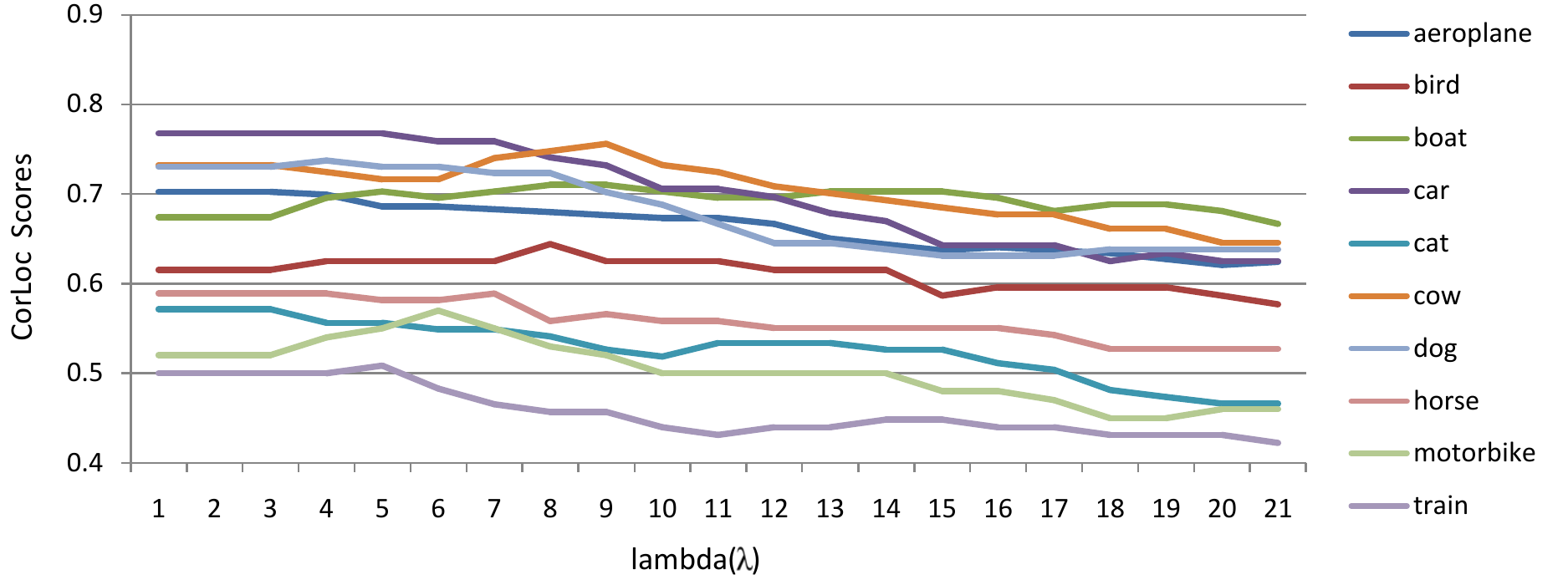}
\end{center}
    \caption{The proposed method's category-wise performance as $\lambda$ varies.}
    \label{fig:lambda}
\end{figure*}
\subsection{Performance variation while varying $\lambda$} Here, we will see how the performance varies with the parameter $\lambda$ introduced in our objective function (5). When varied in steps of one, as shown in Fig.~\ref{fig:lambda}, it can be seen that, for most categories, the performance remains decent for a good range (1-7) but continuously deteriorates later on. As $\lambda$ increases, the model becomes overly dependent upon smoothness and fails to rely on our rough object mask. As a result, the algorithm starts selecting wrong proposals while trying to keep everything smooth. Therefore, we fix our $\lambda$ as 5 in our experiments.

\subsection{Execution Time}
Our algorithm takes about 9 hrs and 17 hrs for co-localizing the objects in the YouTube Objects dataset with sampling rate at one per 10 and 5 frames, respectively. It is comparable to the 16 hrs taken by \cite{Jerripothula2016} and much better than 60 hrs taken by \cite{kwak2015unsupervised}, which has the sampling rate of one per 20 frames.  

\section{Conclusion}
This paper proposes novel approaches to muli-cue fusion and video co-localization. The idea is to fuse the cues in the appearance domain and use hierarchical co-saliency and prior-specific final localization. The proposed approach obtains state-of-the-art co-localization results on YouTube Objects and YouTube Co-localization datasets.

\bibliographystyle{splncs}
\bibliography{main}

\end{document}